\documentclass[11pt,a4paper]{article}
\usepackage[hyperref]{emnlp2020}
\usepackage{times}
\usepackage{latexsym}

\usepackage{microtype}

\usepackage{color,soul}
\setul{0.45ex}{0.25ex}
\usepackage{subcaption}
\usepackage{amsmath}
\usepackage{booktabs}
\usepackage{multirow, makecell}
\usepackage{tikz}
\usepackage{graphicx}
\graphicspath{ {./figures/} }
\usepackage{array}
\usepackage{pgfplots}
\usepackage{algorithm2e}
\usepackage{textcomp}
\usepackage{times}
\usepackage{latexsym}
\usepackage{amsmath}
\usepackage{amsfonts}
\usepackage[noend]{algpseudocode}
\usepackage{lipsum}
\usepackage{multirow}
\usepackage{framed}
\usepackage{graphicx}
\usepackage{pifont}
\usepackage{longtable}
\usepackage{tikz}
\usepackage{booktabs}
\usepackage[all]{nowidow}
\usepackage{enumitem}
\usepackage{color,soul}
\usepackage{amssymb}
\usepackage{colortbl}
\usepackage{tagging}

\newif\ifcomments
\commentstrue 
\ifcomments
    \providecommand{\nascomment}[1]{{\protect\color{red}{[Noah: #1]}}}
    \providecommand{\eric}[1]{{\protect\color{blue}{[Eric: #1]}}}
    \providecommand{\nfliu}[1]{{\protect\color{olive}{[NL: #1]}}}
    \providecommand{\sameer}[1]{{\protect\color{magenta}{[Sameer: #1]}}}
    \providecommand{\matt}[1]{{\protect\color{teal}{[Matt: #1]}}}
    \providecommand{\danielk}[1]{{\protect\color{green}{[DK: #1]}}}
    \providecommand{\gabe}[1]{{\protect\color{purple}{[GI: #1]}}}
    \providecommand{\jb}[1]{{\protect\color{pink}{[JB: #1]}}} 
    
    \providecommand{\sanjay}[1]{{\protect\color{pink}{[Sanjay: #1]}}}
    \providecommand{\hanna}[1]{{\protect\color{cyan}{[Hanna: #1]}}}
    \providecommand{\sihao}[1]{{\protect\color{green}{[Sihao: #1]}}}
    
    \providecommand{\pradeep}[1]{{\protect\color{red}{[Pradeep: #1]}}}
    \providecommand{\qn}[1]{{\protect\color{violet}{[QN: #1]}}}
    \providecommand{\nitish}[1]{{\protect\color{red}{[Nitish: #1]}}}

\else
    \providecommand{\nascomment}[1]{}
    \providecommand{\eric}[1]{}
    \providecommand{\nfliu}[1]{}
    \providecommand{\matt}[1]{}
    \providecommand{\sameer}[1]{}
    \providecommand{\danielk}[1]{}
    \providecommand{\gabe}[1]{}
    \providecommand{\jb}[1]{}
    \providecommand{\yoav}[1]{}
    \providecommand{\sanjay}[1]{}
    \providecommand{\hanna}[1]{}
    \providecommand{\sihao}[1]{}
    \providecommand{\pradeep}[1]{}
    \providecommand{\qn}[1]{}
    \providecommand{\nitish}[1]{}

\fi

\aclfinalcopy % Uncomment this line for the final submission
\newcommand{\numtasks}{10}
\newcommand{\quoref}{\textsc{Quoref}}
\newcommand{\ropes}{\textsc{ROPES}}

\newcommand{\PreserveBackslash}[1]{\let\temp=\\#1\let\\=\temp}
\newcolumntype{C}[1]{>{\PreserveBackslash\centering}p{#1}}
\newcolumntype{R}[1]{>{\PreserveBackslash\raggedleft}p{#1}}
\newcolumntype{L}[1]{>{\PreserveBackslash\raggedright}p{#1}}

\title{Evaluating Models' Local Decision Boundaries via Contrast Sets}
\author{\makecell{Matt Gardner$^{\bigstar \diamondsuit}$ \hfill Yoav Artzi$^\Gamma$ \hfill Victoria Basmova$^{\diamondsuit \clubsuit}$ \hfill Jonathan Berant$^{\diamondsuit \spadesuit}$ \hfill \\
Ben Bogin$^\spadesuit$ \hfill Sihao Chen$^\heartsuit$ \hfill Pradeep Dasigi$^\diamondsuit$ \hfill Dheeru Dua$^\Box$ \hfill Yanai Elazar$^{\diamondsuit \clubsuit}$ \\ Ananth Gottumukkala$^\Box$ \hfill Nitish Gupta$^\heartsuit$ \hfill Hanna Hajishirzi$^{\diamondsuit \triangle}$ \hfill Gabriel Ilharco$^\triangle$ \\ Daniel Khashabi$^\diamondsuit$ \hfill
Kevin Lin$^+$ \hfill Jiangming Liu$^{\diamondsuit \dagger}$ \hfill Nelson F. Liu$^\mathparagraph$ \\ 
Phoebe Mulcaire$^\triangle$ \hfill Qiang Ning$^\diamondsuit$ \hfill Sameer Singh$^\Box$ \hfill Noah A. Smith$^{\diamondsuit \triangle}$ \\ Sanjay Subramanian$^\diamondsuit$ \hfill Reut Tsarfaty$^{\diamondsuit \clubsuit}$ \hfill Eric Wallace$^+$  \hfill Ally Zhang$^\Gamma$ \hfill Ben Zhou$^\heartsuit$} \\ 
$^\diamondsuit$Allen Institute for AI \hfill $^\Gamma$Cornell University \hfill $^\clubsuit$Bar-Ilan University \\ $^\spadesuit$Tel-Aviv University \hfill $^\heartsuit$University of Pennsylvania \hfill $^\triangle$University of Washington \\
$^\Box$UC Irvine  \hfill $^+$UC Berkeley \hfill
$^\dagger$University of Edinburgh  \hfill $^\mathparagraph$Stanford University \\
\href{mailto:mattg@allenai.org}{\tt mattg@allenai.org}}
\date{}

\begin{document}

\maketitle

\begin{abstract}
Standard test sets for supervised learning evaluate in-distribution generalization. Unfortunately,
  when a dataset has systematic gaps (e.g., annotation artifacts), these evaluations are misleading:
  a model can learn simple decision rules that perform well on the test set but do not capture
  the abilities a dataset is intended to test. We propose a more rigorous annotation paradigm for NLP that helps to close
  systematic gaps in the test data. In particular, after a dataset is constructed, we recommend that
  the dataset authors manually perturb the test instances in small but meaningful ways that
  (typically) change the gold label, creating \emph{contrast sets}. Contrast sets provide a local
  view of a model's decision boundary, which can be used to more accurately evaluate a model's true
  linguistic capabilities.  We demonstrate the efficacy of contrast sets by creating them for
  \numtasks{} diverse NLP datasets (e.g., DROP reading comprehension, UD parsing, and IMDb sentiment
  analysis).  Although our contrast sets are not explicitly adversarial, model performance is
  significantly lower on them than on the original test sets---up to 25\% in some cases. We release
  our contrast sets as new evaluation benchmarks and encourage future dataset construction efforts
  to follow similar annotation processes.
\end{abstract}

\section{Introduction}

\newcommand\blfootnote[1]{%
  \begingroup
  \renewcommand\thefootnote{}\footnote{#1}%
  \addtocounter{footnote}{-1}%
  \endgroup
}

\iftagged{anonymized}{}{
\blfootnote{$^\bigstar$ Matt Gardner led the project. All other authors are listed in alphabetical order.}
}

\begin{figure}[!ht]
    \centering
    \includegraphics[width=0.85\columnwidth]{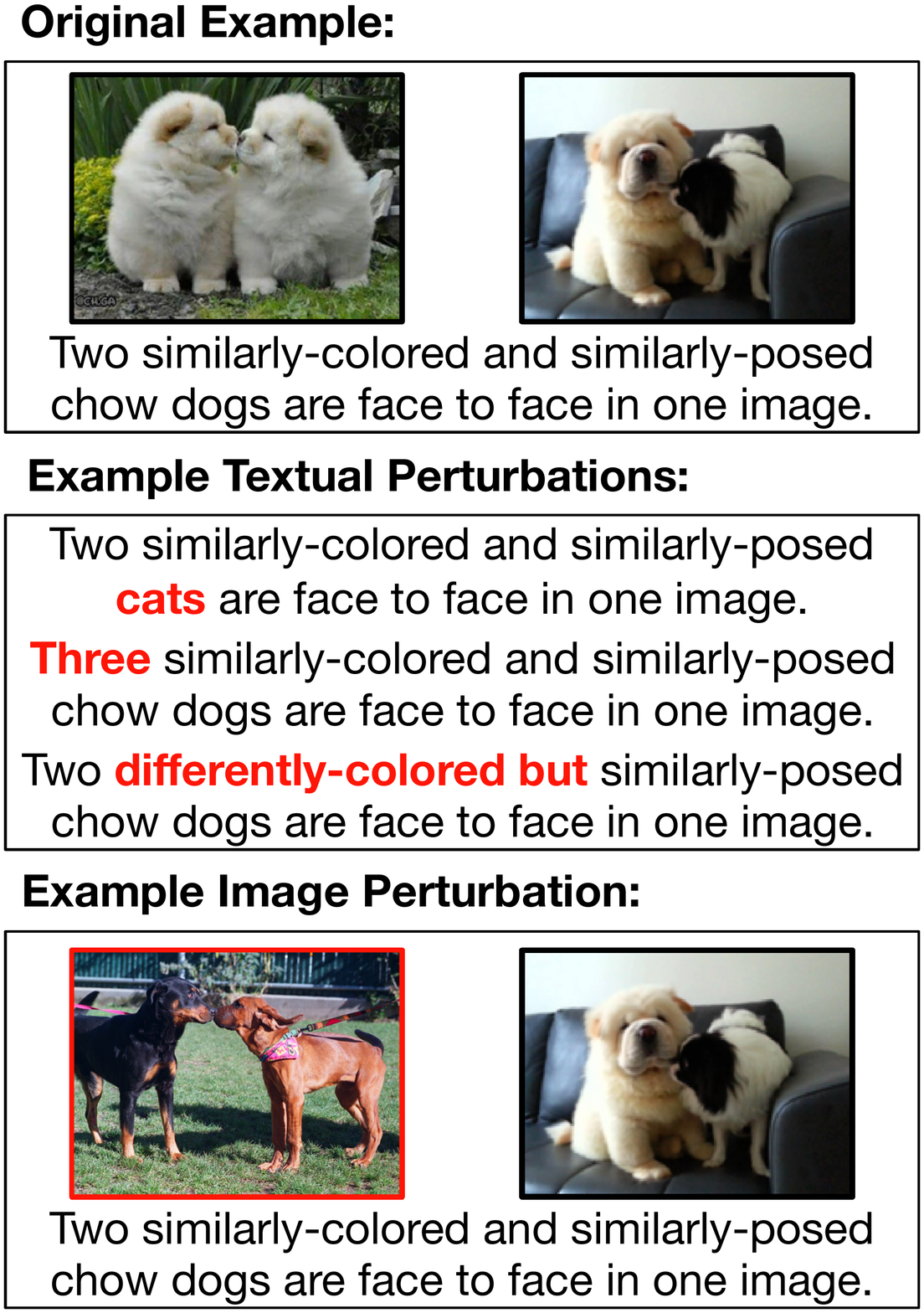}
    \caption{An example contrast set for NLVR2~\cite{suhr2019nlvr2}. The label for the original
    example is \textsc{True} and the label for all of the perturbed examples is \textsc{False}. The
    contrast set allows probing of a model's decision boundary local to examples in the test set, which better evaluates whether
    the model has captured the relevant phenomena than standard metrics on \emph{i.i.d.}~test data.}
    \label{fig:teaser}
\end{figure}

Progress in natural language processing (NLP) has long been measured with standard benchmark
datasets~\cite[e.g.,][]{marcus1993building}.  These benchmarks help to provide a uniform evaluation
of new modeling developments.  However, recent work shows a problem with this standard evaluation
paradigm based on \emph{i.i.d.}~test sets: datasets often have systematic gaps (such as those due to
various kinds of annotator bias) that (unintentionally) allow simple decision rules to perform well
on test data~\cite{chen2016thorough,gururangan2018annotation,geva2019we}.  This is strikingly
evident when models achieve high test accuracy but fail on simple input
perturbations~\cite{jia2017adversarial,feng2018pathologies,ribeiro-etal-2018-semantically}, challenge
examples~\cite{naik2018stress}, and covariate and label
shifts~\cite{ben2010theory,shimodaira2000improving,lipton2018detecting}.

To more accurately evaluate a model's true capabilities on some task, we must collect data that
fills in these systematic gaps in the test set. To accomplish this, we expand on long-standing ideas of constructing minimally-constrastive examples~\cite[e.g.][]{levesque2011winograd}.  We propose that dataset authors
manually perturb instances from their test set, creating \emph{contrast sets} which characterize the correct decision boundary near the test instances (Section~\ref{sec:contrast}).  Following the
dataset construction process, one should make small but (typically) label-changing modifications to
the existing test instances (e.g., Figure \ref{fig:teaser}).  These perturbations should be small,
so that they preserve whatever lexical/syntactic artifacts are present in the original example, but
change the true label.  They should be created \emph{without} a model in the loop, so as not to bias
the contrast sets towards quirks of particular models.  Having a set of contrasting perturbations
for test instances allows for a \emph{consistency} metric that measures how well a model's decision
boundary aligns with the ``correct'' decision boundary around each test instance.

Perturbed test sets only need to be large enough to draw substantiated conclusions about model
behavior and thus do not require undue labor on the original dataset authors.  We show that using
about a person-week of work can yield high-quality perturbed test sets of approximately 1000
instances for many commonly studied NLP benchmarks, though the amount of work varies greatly (Section~\ref{sec:examples}).

We apply this annotation paradigm to a diverse set of \numtasks{} existing NLP datasets---including
visual reasoning, reading comprehension, sentiment analysis, and syntactic parsing---to demonstrate
its wide applicability and efficacy (Section~\ref{sec:results}).  Although contrast sets are not
intentionally adversarial, state-of-the-art models perform dramatically worse on our contrast sets
than on the original test sets, especially when evaluating consistency.
We believe that contrast sets provide a more accurate reflection of a model's true performance, and
we release our datasets as new benchmarks.\footnote{All of our new test sets are available at
\iftagged{anonymized}{\url{https://anon.ymo.us}.}{\url{https://allennlp.org/contrast-sets}.}} We
recommend that creating contrast sets become standard practice for NLP datasets.

\section{Contrast Sets}\label{sec:contrast}

\begin{figure}
    \centering
    
    \begin{subfigure}[b]{0.9\columnwidth}
        \includegraphics[width=\linewidth]{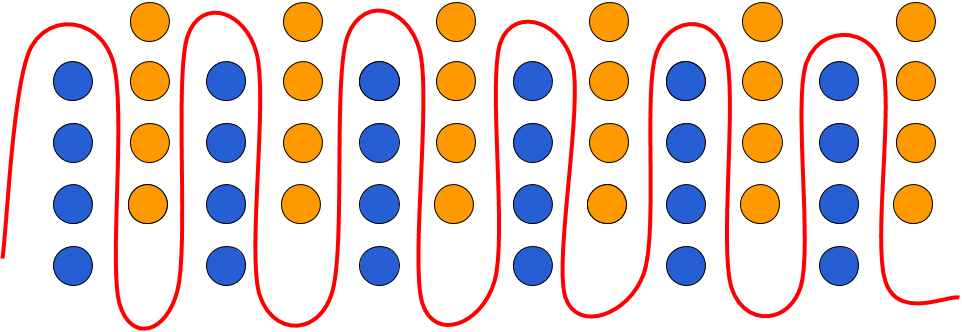}
        \caption{A two-dimensional dataset that requires a complex decision boundary to achieve high accuracy.}
        \label{subfig:dense}
    \end{subfigure}
    \begin{subfigure}[b]{0.9\columnwidth}
        \vspace{5pt}
        \includegraphics[width=\linewidth]{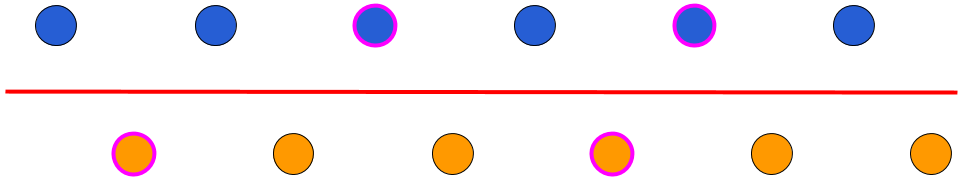}
        \caption{If the same data distribution is instead sampled with systematic gaps (e.g., due to annotator bias), a simple decision boundary \emph{can perform well on i.i.d. test data} (shown outlined in pink).}
        \label{subfig:gaps}
    \end{subfigure}
    \begin{subfigure}[b]{0.9\columnwidth}
        \vspace{5pt}
        \includegraphics[width=\linewidth]{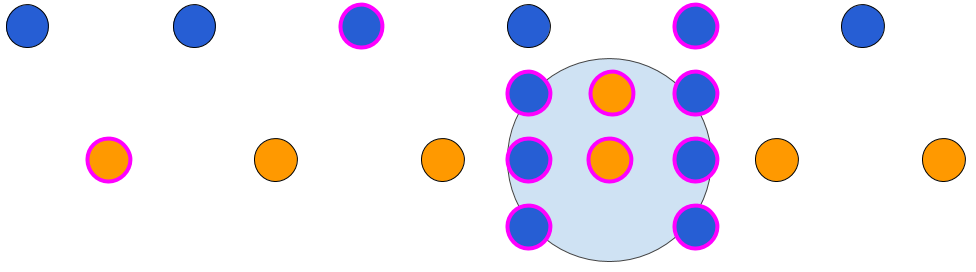}
        \caption{Since filling in all gaps in the distribution is infeasible, a \emph{contrast set} instead fills in a local ball around a test instance to evaluate the model's decision boundary.}
        \label{subfig:contrast}
    \end{subfigure}
    
    \caption{An illustration of how contrast sets provide a more comprehensive model evaluation when datasets have systematic gaps.}
    \label{fig:toy-example}
\end{figure}

\definecolor{adversarial}{rgb}{0.90, 0.02, 0.03}

\begin{table*}[!t]
\centering
\footnotesize
\begin{tabular}{C{2cm}p{6.5cm}p{6.0cm}}
\toprule
{\bf Dataset} & {\bf Original Instance} & {\bf Contrastive Instance} (color = edit)\\
\midrule
\arrayrulecolor{black!30}

\vspace{1.2cm} IMDb & Hardly one to be faulted for his ambition or his vision, it is genuinely unexpected, then, to see all Park's effort add up to so very little. \dots The premise is promising, gags are copious and offbeat humour abounds but it all fails miserably to create any meaningful connection with the audience. \newline {\em (Label: Negative)} &  Hardly one to be faulted for his ambition or his vision, \textbf{\textcolor{blue}{here we see all Park's effort come to fruition.}} \dots The premise is \textbf{\textcolor{blue}{perfect}}, gags are \textbf{\textcolor{blue}{hilarious}} and offbeat humour abounds, \textbf{\textcolor{blue}{and it creates a deep}} connection with the audience. \newline {\em (Label: Positive)} \\

\midrule

\vspace{0.1cm} MATRES 
& Colonel Collins followed a normal progression once she was picked as a NASA astronaut. \newline {\em (``picked'' was before ``followed'')} 
& Colonel Collins followed a normal progression \textbf{\color{olive} before} she was picked as a NASA astronaut. \newline {\em (``picked'' was after ``followed'')} \\

\midrule

UD English &
\begin{minipage}{6.2cm}
They demanded talks with local US commanders.\vspace{0.1cm}\\
I attach a paper on gas storage value modeling. \vspace{0.1cm}\\
I need to get a job at the earliest opportunity.
\end{minipage}
&
\begin{minipage}{6.2cm}
They demanded talks with \textbf{\color{magenta} great urgency}.\vspace{0.1cm} \\
I attach a paper on \textbf{\color{magenta} my own initiative}.\vspace{0.1cm} \\
I need to get a job at \textbf{\color{magenta} House of Pies}.
\end{minipage} \\
\midrule

PERSPECTRUM & 
\begin{minipage}{6.2cm} 
\textbf{Claim:} Should uniforms be worn at school. \\ 
\textbf{Perspective:} School uniforms emphasize the socio-economic divisions they are supposed to eliminate. \\
\textbf{Label:} Against 
\end{minipage} 
  & 
\begin{minipage}{6.2cm} 
\textbf{Claim:} Should uniforms be \textbf{\color{adversarial} banned} at school. \\ 
\textbf{Perspective:} School uniforms emphasize the socio-economic divisions they are supposed to eliminate. \\
\textbf{Label:} For 
\end{minipage}  \\

\midrule
DROP &
\begin{minipage}{6.3cm}
\textbf{Context:} In the spring of 1625 the Spanish regained Bahia in Brazil and Breda in the Netherlands from the Dutch. In the autumn they repulsed the English at Cadiz. \\
\textbf{Question:} What event happened first, the Spanish repulsed the English at Cadiz or the Spanish regained Bahia?
\end{minipage} & 

\begin{minipage}{6.3cm}
\textbf{Context:} In the spring of 1625 the Spanish regained Bahia in Brazil and Breda in the Netherlands from the Dutch. In \textbf{\color{teal} winter the year earlier} they had repulsed the English at Cadiz. \\
\textbf{Question:} What event happened first, the Spanish repulsed the English at Cadiz or the Spanish regained Bahia?
\end{minipage}\\
\midrule

\quoref & \begin{minipage}{6.2cm} \textbf{Context:} Matt Helm is a secret agent. His assignment is to stop the sinister Tung-Tze, armed with spy gadgets. Helm prevails with Gail by his side as he destroys Tung-Tze.\\ \textbf{Question:} Who is armed with spy gadgets?  \end{minipage}& \begin{minipage}{6.2cm}\textbf{Context:} Matt Helm is a secret agent. His assignment is to stop the sinister Tung-Tze, \textbf{\color{violet} even though he is} armed with spy gadgets. Helm prevails with Gail by his side as he destroys Tung-Tze.\\  \textbf{Question:} Who is armed with spy gadgets? \end{minipage}\\
\midrule

MC-TACO & 
\begin{minipage}{6.2cm} 
\textbf{Context:} She renews in Ranchipur an acquaintance with a former lover, Tom Ransome, now a dissolute alcoholic. \\ 
\textbf{Question:} How frequently does Tom drink? \\ 
\textbf{Candidate Answer:} Every other night \\
\textbf{Label:} Likely 
\end{minipage}
& 
\begin{minipage}{6.2cm} 
\textbf{Context:} She renews in Ranchipur an acquaintance with a former lover, Tom Ransome, \textbf{\color{orange}who keeps very healthy habits}. \\ 
\textbf{Question:} How frequently does Tom drink? \\ 
\textbf{Candidate Answer:} Every other night \\
\textbf{Label:} Unlikely 
\end{minipage}
 \\
\arrayrulecolor{black}
\bottomrule
\end{tabular}
\caption{We create contrast sets for \numtasks{} datasets and show instances from seven of them here.}
\label{table:examples}
\end{table*}

\subsection{The Problem}

We first give a sketch of the problem that contrast sets attempt to solve in a toy two-dimensional classification setting as shown in
Figure~\ref{fig:toy-example}. Here, the true underlying data distribution requires a complex
decision boundary (Figure~\ref{subfig:dense}).  However, as is common in practice, our toy dataset
is rife with systematic gaps (e.g., due to annotator bias, repeated patterns, etc.). This causes
simple decision boundaries to emerge (Figure~\ref{subfig:gaps}). And, because our biased dataset is
split \emph{i.i.d.}~into train and test sets, this simple decision boundary will perform well on
test data. Ideally, we would like to fill in all of a dataset's systematic gaps, however, this is
usually impossible. Instead, we create a \emph{contrast set}: a collection of instances tightly
clustered in input space around a single test instance, or \emph{pivot}
(Figure~\ref{subfig:contrast}; an $\epsilon$-ball in our toy example).  This contrast set allows us
to measure how well a model's decision boundary aligns with the correct decision boundary local to
the pivot. In this case, the contrast set demonstrates that the model's simple decision boundary is
incorrect. We repeat this process around numerous pivots to form
entire evaluation datasets.

When we move from toy settings to complex NLP tasks, the precise nature of a ``systematic gap'' in
the data becomes harder to define. Indeed, the geometric view in our toy examples does not correspond directly to experts' perception of data; there are many ways to ``locally perturb'' natural language.  We do not expect intuition, even of experts, to exhaustively reveal gaps.

Nevertheless, the presence of these gaps is
well-documented~\cite{gururangan2018annotation,poliak2018hypothesis,min2019compositional}, and \citet{niven-kao-2019-probing} give an initial attempt at formally characterizing them.
In particular, one common source is annotator bias from data collection
processes~\cite{geva2019we}. For example, in the SNLI dataset~\cite{bowman2015large},
\citet{gururangan2018annotation} show that the words \emph{sleeping}, \emph{tv}, and \emph{cat}
almost never appear in an entailment example, either in the training set or the test set, though
they often appear in contradiction examples.  This is not because these words are particularly
important to the phenomenon of entailment; their absence in entailment examples is a
\emph{systematic gap} in the data that can be exploited by models to achieve artificially high test
accuracy.  This is but one kind of systematic gap; there are also biases due to the
writing styles of small groups of annotators~\cite{geva2019we}, the distributional biases in
the data that was chosen for annotation, as well as numerous other biases that are more subtle and
harder to discern~\cite{Shah2019PredictiveBI}.

Completely removing these gaps in the initial data collection process would be ideal, but is likely
impossible---language has too much inherent variability in a very high-dimensional space.  Instead, we use contrast sets to fill in gaps in the test data to give more thorough
evaluations than what the original data provides.

\subsection{Definitions}

We begin by defining a {\bf \emph{decision boundary}} as a partition of some space into labels.\footnote{In this discussion we are talking about the \emph{true} decision boundary, not a \emph{model's} decision boundary.} This partition can be represented by the set of all points in the space with their associated labels: $\{(x, y)\}$.  This definition differs somewhat from the canonical definition, which is a collection of hypersurfaces that separate labels.  There is a bijection between partitions and these sets of hypersurfaces in continuous spaces, however, so they are equivalent definitions.  We choose to use the partition to represent the decision boundary as it makes it very easy to define a \emph{local} decision boundary and to generalize the notion to discrete spaces, which we deal with in NLP.

A {\bf \emph{local decision boundary}} around some {\bf \emph{pivot}} $x$ is the set of all points $x'$ and their associated labels $y'$ that are within some distance $\epsilon$ of $x$. That is, a local decision boundary around $x$ is the set $\{(x', y')~|~d(x, x') < \epsilon\}$.  Note here that even though a ``boundary'' or ``surface'' is hard to visualize in a discrete input space, using this partition representation instead of hypersurfaces gives us a uniform definition of a local decision boundary in any input space; all that is needed is a distance function $d$.

A {\bf \emph{contrast set}} $C(x)$ is any sample of points from a local decision boundary around $x$. In other words, $C(x)$ consists of inputs $x'$ that are similar to $x$ according to some distance function $d$. Typically these points are sampled such that $y' \ne y$. To evaluate a model using these contrast sets, we define the {\bf \emph{contrast consistency}} of a model to be whether it makes correct predictions $\hat{y}$ on every element in the set: $\mathrm{all}(\{\hat{y} = y'~\forall (x', y') \in C(x)\})$. Since the points $x'$ were chosen from the local decision boundary, we expect contrast consistency on expert-built contrast sets to be a significantly more accurate evaluation of whether model predictions match the task definition than a random selection of input / output pairs.

\subsection{Contrast sets in practice}

Given these definitions, we now turn to the actual construction of contrast sets in practical NLP settings.  There were two things left unspecified in the definitions above: the distance function $d$ to use in discrete input spaces, and the method for sampling from a local decision boundary.  While there has been some work trying to formally characterize distances for adversarial robustness in NLP~\cite{michel2019adversarial,jia-etal-2019-certified}, we find it more useful in our setting to simply rely on expert judgments to generate a similar but meaningfully different $x'$ given $x$, addressing both the distance function and the sampling method.

Future work could try to give formal treatments of these issues, but we believe expert judgments are sufficient to make initial progress in improving our evaluation methodologies.  And while expert-crafted contrast sets can only give us an upper bound on a model's local alignment with the true decision boundary, an upper bound on local alignment is often more informative than a potentially biased \emph{i.i.d.} evaluation that permits artificially simple decision boundaries.  To give a tighter upper bound, we draw pivots $x$ from some \emph{i.i.d.} test set, and we do not provide \emph{i.i.d.} contrast sets at training time, which could provide additional artificially simple decision boundaries to a model.

Figure \ref{fig:teaser} displays an example contrast set for the NLVR2 visual
reasoning dataset~\cite{suhr2019nlvr2}.  Here, both the sentence and the image are modified in small
ways (e.g., by changing a word in the sentence or finding a similar but different image) to make the
output label change.

A contrast set is \emph{not} a collection of adversarial examples~\cite{szegedy2013-intriguing}.
Adversarial examples are almost the methodological opposite of contrast sets: they change the input
such that a model's decision changes but the gold label does
not~\cite{jia2017adversarial,wallace2019triggers}. On the other hand, contrast sets are model-agnostic, constructed by experts to characterize whether a model's decision boundary locally aligns to the true decision boundary around some point.  Doing this requires input changes that also induce changes to the gold label.

We recommend that the original dataset authors---the experts on the linguistic phenomena intended to be reflected in 
their dataset---construct the contrast sets. This is best done by first identifying a list of
phenomena that characterize their dataset. In syntactic parsing, for example, this list might
include prepositional phrase attachment ambiguities, coordination scope, clausal attachment, etc.
After the standard dataset collection process, the authors should sample pivots from their test set
and perturb them according to the listed phenomena.

\subsection{Design Choices of Contrast Sets}
\label{subsec:design_choices}

Here, we discuss possible alternatives to our approach for constructing contrast sets and our
reasons for choosing the process we did.

\iftagged{anonymized}{}{
\paragraph{Post-hoc Construction of Contrast Sets} Improving the evaluation for existing datasets
well after their release is usually too late: new models have been designed, research papers have
been published, and the community has absorbed potentially incorrect insights. Furthermore, post-hoc
contrast sets may be biased by existing models. We instead recommend that new datasets include
contrast sets upon release, so that the authors can characterize beforehand when they will be
satisfied that a model has acquired the dataset's intended capabilities. Nevertheless, contrast sets
constructed post-hoc are still better than typical \emph{i.i.d.}~test sets, and where feasible we
recommend creating contrast sets for existing datasets (as we do in this work).
}

\iftagged{anonymized}{}{
\paragraph{Crowdsourcing Contrast Sets} We recommend that the dataset authors construct contrast
sets themselves rather than using crowd workers. The original authors are the ones who best
understand their dataset's intended phenomena and the distinction between in-distribution and
out-of-distribution examples---these ideas can be difficult to distill to non-expert crowd workers.
Moreover, the effort to create contrast sets is a small fraction of the effort required to produce a
new dataset in the first place.
}

\paragraph{Automatic Construction of Contrast Sets} Automatic perturbations, such as paraphrasing
with back-translation or applying word replacement rules, can fill in some parts of the gaps around
a pivot~\cite[e.g.,][]{ribeiro2018semantically,ribeiro-etal-2019-red}. However, it is very
challenging to come up with rules or other automated methods for pushing pivots \emph{across a
decision boundary}---in most cases this presupposes a model that can already perform the intended
task. We recommend annotators spend their time constructing these types of examples; easier examples
can be automated.

\paragraph{Adversarial Construction of Contrast Sets} Some recent datasets are constructed using
baseline models in the data collection process, either to filter out examples that existing models
answer correctly~\cite[e.g.,][]{dua2019drop,dasigi2019quoref} or to generate adversarial inputs
\cite[e.g.,][]{zellers2018swag,zellers2019hellaswag,wallace2019trick,nie2019adversarial}. Unlike
this line of work, we choose \textit{not} to have a model in the loop because this can bias the data
to the failures of a particular model~\cite[cf.][]{zellers2019hellaswag}, rather than generally
characterizing the local decision boundary. We do think it is acceptable to use a model on
a handful of initial perturbations to understand which phenomena are worth spending time on, but
this should be separate from the actual annotation process---observing model outputs while
perturbing data creates subtle, undesirable biases towards the idiosyncrasies of that model.
 
\subsection{Limitations of Contrast Sets}

\paragraph{Solely Negative Predictive Power} Contrast sets only have negative predictive power: they
reveal if a model \emph{does not} align with the correct local decision boundary but cannot confirm
that a model \emph{does} align with it. This is because annotators cannot exhaustively label all
inputs near a pivot and thus a contrast set will necessarily be incomplete. However, note that this
problem is not unique to contrast sets---similar issues hold for the original test set as well as
adversarial test sets~\cite{jia2017adversarial}, challenge sets~\cite{naik2018stress}, and input
perturbations~\cite{ribeiro-etal-2018-semantically,feng2018pathologies}. \iftagged{anonymized}{}{See
\citet{feng2019misleading} for a detailed discussion of how dataset analysis methods only have
negative predictive power.}

\paragraph{Dataset-Specific Instantiations} The process for creating contrast sets is
\textit{dataset-specific}: although we present general guidelines that hold across many tasks,
experts must still characterize the type of phenomena each individual dataset is intended to
capture. Fortunately, the original dataset authors should \textit{already} have thought deeply about
such phenomena. Hence, creating contrast sets should be well-defined and relatively straightforward.

\section{How to Create Contrast Sets}\label{sec:examples}

Here, we walk through our process for creating contrast sets for three datasets. Examples are shown in Figure~\ref{fig:teaser} and Table~\ref{table:examples}.

\paragraph{DROP} DROP~\cite{dua2019drop} is a reading comprehension dataset that is intended to
cover compositional reasoning over numbers in a paragraph, including filtering, sorting, and
counting sets, and doing numerical arithmetic. The data has three main sources of paragraphs, all
from Wikipedia articles: descriptions of American football games, descriptions of census results,
and summaries of wars.  There are many common patterns used by the crowd workers that make some
questions artificially easy: 2 is the most frequent answer to \emph{How many\dots?} questions,
questions asking about the ordering of events typically follow the linear order of the paragraph,
and a large fraction of the questions do not require compositional reasoning.

Our strategy for constructing contrast sets for DROP was three-fold. First, we added more
compositional reasoning steps.  The questions about American football passages in the original data
very often had multiple reasoning steps (e.g., \emph{How many yards difference was there between the
Broncos' first touchdown and their last?}), but the questions about the other passage types did not.
We drew from common patterns in the training data and added additional reasoning steps to questions
in our contrast sets. Second, we inverted the semantics of various parts of the question.  This
includes perturbations such as changing \emph{shortest} to \emph{longest}, \emph{later} to
\emph{earlier}, as well as changing questions asking for counts to questions asking for sets
(\emph{How many countries\dots} to \emph{Which countries\dots}). Finally, we changed the ordering of
events. A large number of questions about war paragraphs ask which of two events happened first.  We
changed (1) the order the events were asked about in the question, (2) the order that the events
showed up in the passage, and (3) the dates associated with each event to swap their temporal order.

\paragraph{NLVR2} We next consider NLVR2, a dataset where a model is given a sentence about two
provided images and must determine whether the sentence is true~\cite{suhr2018corpus}. The data
collection process encouraged highly compositional language, which was intended to require
understanding the relationships between objects, properties of objects, and counting. We constructed
NLVR2 contrast sets by modifying the sentence or replacing one of the images with freely-licensed
images from web searches.  For example, we might change \emph{The left image contains twice the
number of dogs as the right image} to \emph{The left image contains \textbf{three times} the number
of dogs as the right image}. Similarly, given an image pair with four dogs in the left and two dogs
in the right, we can replace individual images with photos of variably-sized groups of dogs. The
textual perturbations were often changes in quantifiers (e.g., \emph{at least one} to \emph{exactly
one}), entities (e.g., \emph{dogs} to \emph{cats}), or properties thereof (e.g., \emph{orange glass}
to \emph{green glass}).  An example contrast set for NLVR2 is shown in Figure~\ref{fig:teaser}.

\paragraph{UD Parsing} Finally, we discuss dependency parsing in the universal dependencies (UD)
formalism~\cite{ud}. We look at dependency parsing to show that contrast sets apply not only to
modern ``high-level'' NLP tasks but also to longstanding linguistic analysis tasks.  We first chose
a specific type of attachment ambiguity to target: the classic problem of prepositional phrase (PP)
attachment~\cite{collins-brooks-1995-prepositional}, e.g. \emph{We ate spaghetti with forks} versus
\emph{We ate spaghetti with meatballs}. We use a subset of the English UD treebanks:
GUM~\cite{Zeldes2017}, the English portion of LinES~\cite{ahrenberg2007lines}, the English portion
of ParTUT~\cite{sanguinetti2015parttut}, and the dependency-annotated English Web
Treebank~\cite{silveira2014gold}. We searched these treebanks for sentences that include a
potentially structurally ambiguous attachment from the head of a PP to either a
noun or a verb. We then perturbed these sentences by altering one of their noun phrases such that
the semantics of the perturbed sentence required a different attachment for the PP. We then re-annotated these perturbed sentences to indicate the new attachment(s).

\begin{table*}[tbh]
\small
\centering
\begin{tabular}{lrr|lrrrr}
    \toprule
    \textbf{Dataset} & \textbf{\# Examples} & \textbf{\# Sets} & \textbf{Model} & \textbf{Original Test}  & \multicolumn{2}{c}{\textbf{Contrast}} & \textbf{Consistency} \\
    \midrule

NLVR2 & 994 & 479 & LXMERT & 76.4 & 61.1 & (--15.3) & 30.1 \\[3mm]

IMDb & 488 & 488 & BERT & 93.8 & 84.2 &(--9.6) & 77.8 \\[3mm]

MATRES & 401 & 239 & CogCompTime2.0 &  73.2 & 63.3 & (--9.9) & 40.6 \\[3mm]

UD English & 150 & 150 & Biaffine + ELMo & 64.7 & 46.0 &(--18.7) & 17.3 \\[3mm]

PERSPECTRUM & 217 & 217 & RoBERTa & 90.3 & 85.7 & (--4.6) & 78.8 \\[3mm]

DROP & 947 & 623 & MTMSN & 79.9 & 54.2 & (--25.7) & 39.0 \\[3mm]

QUOREF & 700 & 415 & XLNet-QA & 70.5 & 55.4 & (--15.1) & 29.9 \\[3mm]

ROPES & 974 & 974 & RoBERTa & 47.7 & 32.5 & (--15.2) & 17.6 \\[3mm]

BoolQ & 339 & 70 & RoBERTa & 86.1 & 71.1 & (--15.0) & 59.0  \\[3mm]

MC-TACO & 646 & 646 & RoBERTa & 38.0 & 14.0 & (--24.0) & 8.0 \\
\bottomrule
\end{tabular}
    \caption{Models struggle on the contrast sets compared to the original test sets. For each
    dataset, we use a (sometimes near) state-of-the-art model and evaluate it on the
    ``\# Examples'' examples in the contrast sets (\emph{not} including the original example). We
    report percentage accuracy for NLVR2, IMDb, PERSPECTRUM, MATRES, and BoolQ; F$_1$ scores for
    DROP and \quoref; Exact Match (EM) scores for ROPES and MC-TACO; and unlabeled attachment score
    on modified attachments for the UD English dataset. We also report \emph{contrast consistency}:
    the percentage of the ``\# Sets'' contrast sets for which a model's predictions are correct for
    all examples in the set (\emph{including} the original example). More details on datasets,
    models, and metrics can be found in \S\ref{appendix:dataset_details} and
    \S\ref{appendix:contrast-set-details}.}
    \label{tab:results}
\end{table*}

 % put the table here so it is displayed in a good spot in the paper.

\paragraph{Summary} While the overall process we recommend for constructing contrast sets is simple
and unified, its actual instantiation varies for each dataset. Dataset authors should use their best
judgment to select which phenomena they are most interested in studying and craft their contrast
sets to explicitly test those phenomena.  Care should be taken during contrast set construction to ensure that the phenomena present in contrast sets are similar to those present in the original test set; the purpose of a contrast set is not to introduce new challenges, but to more thoroughly evaluate the original intent of the test set.

\section{Datasets and Experiments}\label{sec:results}

\subsection{Original Datasets}\label{subsec:datasets}

We create contrast sets for \numtasks{} NLP datasets (full descriptions are provided in
Section~\ref{appendix:dataset_details}):
\begin{itemize}[nosep,leftmargin=6mm]
    \item \textbf{NLVR2}~\cite{suhr2018corpus}
    \item \textbf{IMDb sentiment analysis}~\cite{maas2011imdb}
    \item \textbf{MATRES Temporal RE}~\cite{ning2018matres}
    \item \textbf{English UD parsing}~\cite{ud}
    \item \textbf{PERSPECTRUM}~\cite{chen2019seeing}
    \item \textbf{DROP}~\cite{dua2019drop}
    \item \textbf{Quoref}~\cite{dasigi2019quoref}
    \item \textbf{ROPES}~\cite{lin2019reasoning}
    \item \textbf{BoolQ}~\cite{clark2019boolq}
    \item \textbf{MC-TACO}~\cite{zhou2019going}
\end{itemize}

We choose these datasets because they span a variety of tasks (e.g., reading comprehension,
sentiment analysis, visual reasoning) and input-output formats (e.g., classification, span
extraction, structured prediction).  We include high-level tasks for which dataset artifacts are
known to be prevalent, as well as longstanding formalism-based tasks, where data artifacts have been
less of an issue (or at least have been less well-studied).

\subsection{Contrast Set Construction}

The contrast sets were constructed by NLP researchers who were deeply familiar with the phenomena
underlying the annotated dataset\iftagged{anonymized}{.}{; in most cases, these were the original
dataset authors.} Our contrast sets consist of up to about 1,000 total examples and average 1--5
examples per contrast set (Table~\ref{tab:results}). We show representative examples from the
different contrast sets in Table~\ref{table:examples}. For most datasets, the average time to
perturb each example was 1--3 minutes, which translates to approximately 17--50 hours of work to
create 1,000 examples. However, some datasets, particularly those with complex output structures,
took substantially longer: each example for dependency parsing took an average of 15 minutes (see
Appendix~\ref{appendix:contrast-set-details} for more details).

\subsection{Models Struggle on Contrast Sets}

For each dataset, we use a model that is at or near state-of-the-art performance. Most models
involve fine-tuning a pretrained language model (e.g., ELMo~\cite{PetersELMo2018}, BERT~\cite{devlin2018BERT},
\textsc{RoBERTa}~\cite{liu2019roberta}, XLNet~\cite{yang2019xlnet}, etc.) or applying a
task-specific architecture on top of one (e.g., \citet{hu2019mtmsn} add a DROP-specific model on top
of BERT). We train each model on the original training set and evaluate it on both the original test
set and our contrast sets.

Existing models struggle on the contrast sets (Table~\ref{tab:results}), particularly when
evaluating contrast consistency. Model performance degrades differently across datasets; however,
note that these numbers are not directly comparable due to differences in dataset size, model
architecture, contrast set design, etc.  On IMDb and PERSPECTRUM, the model achieves a reasonably
high consistency, suggesting that, while there is definitely still room for improvement, the
phenomena targeted by those datasets are already relatively well captured by existing models.

Of particular note is the very low consistency score for dependency parsing.  The parser that
we use achieves 95.7\% unlabeled attachment score on the English Penn Treebank~\cite[][trained with ELMo embeddings]{dozat2016deep}.
A consistency score of 17.3 on a common attachment ambiguity suggests that this parser
may not be as strong as common evaluations lead us to believe.  Overall, our results suggest that
models have ``overfit'' to artifacts that are present in existing datasets; they achieve high test
scores but do not completely capture a dataset's intended phenomena.

\subsection{Humans Succeed On Contrast Sets}\label{subsec:human}

An alternative explanation for why models fail on the contrast sets is that they are
simply harder or noisier than regular test sets, i.e., humans would also perform worse on the
contrast sets. We show that this is not the case. For four datasets, we choose at least 100
test instances and one corresponding contrast set instance (i.e., an example before and
after perturbation). We (the authors) test ourselves on these examples (ensuring that those who were tested were different from those who created the examples). Human performance is
comparable across the original test and contrasts set examples on these datasets
(Table~\ref{tab:human_results}).

\begin{table}[tbh]
    \small
    \centering
    \begin{tabular}{lrrr}
        \toprule
        \textbf{Dataset} & \textbf{Original Test} & \multicolumn{2}{c}{\textbf{Contrast Set}}\\
        \midrule
IMDb & 94.3 & 93.9  & (--0.4) \\
\addlinespace

PERSPECTRUM & 91.5 & 90.3 & (--1.2) \\
\addlinespace

\quoref & 95.2 & 88.4 & (--6.8) \\
\addlinespace

ROPES  & 76.0 & 73.0 & (--3.0) \\
\bottomrule
\end{tabular}
    \caption{Humans achieve similar performance on the contrast sets and the original test sets. The
    metrics here are the same as those in Table~\ref{tab:results}.}
    \label{tab:human_results}
\end{table}

\subsection{Fine-Grained Analysis of Contrast Sets}

Each example in the contrast sets can be labeled according to which particular phenomenon it
targets. This allows automated error reporting. For example, for the MATRES dataset we tracked
whether a perturbation changed appearance order, tense, or temporal conjunction words. These
fine-grained labels show that the model does comparatively better at modeling appearance order
(66.5\% of perturbed examples correct) than temporal conjunction words (60.0\% correct); see
Appendix~\ref{subsec:matres} for full details. A similar analysis on DROP shows that MTMSN does
substantially worse on event re-ordering (47.3 F$_1$) than on adding compositional reasoning steps
(67.5 F$_1$). We recommend authors categorize their perturbations up front in order to
simplify future analyses and bypass some of the pitfalls of post-hoc error
categorization~\cite{wu-etal-2019-errudite}.

Additionally, it's worth discussing the dependency parsing result.  The attachment decision that we
targeted was between a verb, a noun, and a preposition.  With just two reasonable attachment
choices, a contrast consistency of 17.3 means that the model is almost always unable to change its
attachment based on the content of the prepositional phrase.  Essentially, in a trigram such as
\emph{demanded talks with} (Table~\ref{table:examples}), the model has a bias for whether
\emph{demanded} or \emph{talks} has a stronger affinity to \emph{with}, and makes a prediction
accordingly.  Given that trigrams are rare and annotating parse trees is expensive, it is not clear
that traditional evaluation metrics with \emph{i.i.d}~test sets would ever find this problem.  By
robustly characterizing local decision boundaries, contrast sets surface errors that are very
challenging to find with other means.

\section{Related Work}\label{sec:discussion}

The fundamental idea of finding or creating data that is ``minimally different'' has a very long history. In linguistics, for instance, the term \emph{minimal pair} is used to denote two words with different meaning that differ by a single sound change, thus demonstrating that the sound change is phonemic in that language~\cite{pike1946phonemics}. Many people have used this idea in NLP (see below), creating challenge sets or providing training data that is ``minimally different'' in some sense, and we continue this tradition. Our main contribution to this line of work, in addition to the resources that we have created, is giving a simple and intuitive geometric interpretation of ``bias'' in dataset collection, and showing that this long-standing idea of minimal data changes can be effectively used to solve this problem on a wide variety of NLP tasks. We additionally generalize the idea of a minimal \emph{pair} to a \emph{set}, and use a \emph{consistency} metric, which we contend more closely aligns with what NLP researchers mean by ``language understanding''.

\paragraph{Training on Perturbed Examples} 
Many previous works have provided minimally contrastive examples on which to train models. \citet{selsam2018learning}, \citet{tafjord-etal-2019-quartz}, \citet{lin2019reasoning}, and \citet{khashabi2020naturalperturbations} designed their data collection process to include contrastive examples.  Data augmentation methods have also been used to mitigate gender~\cite{zhao2018gender}, racial~\cite{dixon2018measuring}, and other biases~\cite{kaushik2019learning} during training, or to introduce useful inductive biases~\cite{andreas-2020-good}.

\paragraph{Challenge Sets} The idea of creating challenging contrastive evaluation sets has a long history~\cite{levesque2011winograd,ettinger-etal-2017-towards,glockner2018breaking,naik2018stress,isabelle2017challenge}. Challenge sets exist for
various phenomena, including ones with ``minimal'' edits similar to our contrast sets, e.g., in
image captioning~\cite{shekhar2017foil}, machine translation~\cite{sennrich2016grammatical,burlot-yvon-2017-evaluating,burlot-etal-2018-wmt18}, and
language modeling~\cite{marvin2018targeted,warstadt2019blimp}.  Minimal pairs of edits that perturb
gender or racial attributes are also useful for evaluating social
biases~\cite{rudinger2018gender,zhao2018gender,lu2018gender}.  Our key contribution over this prior
work is in grouping perturbed instances into a contrast set, for measuring local alignment of decision
boundaries, along with our new, related resources.  Additionally, rather than creating new data from
scratch, contrast sets augment existing test examples to fill in systematic gaps. Thus contrast sets often require less effort to create, and they remain grounded in the original data distribution of some training set.

Since the initial publication of this paper, \citet{Shmidman2020ChallengeSet} have further demonstrated the utility of contrast sets by applying these ideas to the evaluation of morphological disambiguation in Hebrew.

\iftagged{anonymized}{}{
\paragraph{Recollecting Test Sets} \citet{recht2019imagenet} create new test sets for CIFAR and
ImageNet by closely following the procedure used by the original datasets authors;
\citet{yadav2019cold} perform similar for MNIST. This line of work looks to evaluate whether
\textit{reusing} the exact same test set in numerous research papers causes the community to
adaptively ``overfit'' its techniques to that test set. Our goal with contrast sets is
different---we look to eliminate the biases in the \textit{original annotation process} to better
evaluate models. This cannot be accomplished by simply collecting more data because the new data
will capture similar biases.
}
\section{Conclusion}

We presented a new annotation paradigm, based on long-standing ideas around contrastive examples, for constructing more rigorous test sets for NLP. Our
procedure maintains most of the established processes for dataset creation but fills in some of the
systematic gaps that are typically present in datasets. By shifting evaluations from accuracy on
\emph{i.i.d.}~test sets to consistency on contrast sets, we can better examine whether models have
learned the desired capabilities or simply captured the idiosyncrasies of a dataset. We created
contrast sets for \numtasks{} NLP datasets and released this data as new evaluation benchmarks.

We recommend that future data collection efforts create contrast sets to provide more comprehensive
evaluations for both existing and new NLP datasets. While we have created thousands of new test
examples across a wide variety of datasets, we have only taken small steps towards the rigorous
evaluations we would like to see in NLP. The last several years have given us dramatic modeling
advancements; our evaluation methodologies and datasets need to see similar improvements.

\section*{Acknowledgements}

We thank the anonymous reviewers for their helpful feedback on this paper, as well as many others who gave constructive comments on a publicly-available preprint. Various authors of this paper were supported in part by ERC grant 677352, NSF grant 1562364, NSF grant IIS-1756023, NSF CAREER 1750499, ONR grant N00014-18-1-2826 and DARPA grant N66001-19-2-403.

\bibliography{journal-abbrv,bib}
\bibliographystyle{acl_natbib}

\appendix
\clearpage 

\section{Dataset Details}\label{appendix:dataset_details}

Here, we provide details for the datasets that we build contrast sets for.

\paragraph{Natural Language Visual Reasoning 2} (NLVR$2$) Given a natural language sentence about
two photographs, the task is to determine if the sentence is true~\cite{suhr2018corpus}.  The
dataset has highly compositional language, e.g., \emph{The left image contains twice the number of
dogs as the right image, and at least two dogs in total are standing}. To succeed at NLVR2, a model
is supposed to be able to detect and count objects, recognize spatial relationships, and understand
the natural language that describes these phenomena.

\paragraph{Internet Movie Database} (IMDb) The task is to predict the sentiment (positive or
negative) of a movie review~\cite{maas2011imdb}.  We use the same set of reviews from
\citet{kaushik2019learning} in order to analyze the differences between crowd-edited reviews and
expert-edited reviews.

\paragraph{Temporal relation extraction} (MATRES) The task is to determine what temporal
relationship exists between two events, i.e., whether some event happened \emph{before} or
\emph{after} another event~\cite{ning2018matres}. MATRES has events and temporal relations labeled
for approximately 300 news articles. The event annotations are taken from the data provided in the
TempEval3 workshop~\cite{uzzaman-etal-2013-semeval} and the temporal relations are re-annotated
based on a multi-axis formalism. We assume that the events are given and only need to classify the
relation label between them.

\paragraph{English UD Parsing} We use a combination of four English treebanks (GUM, EWT, LinES,
ParTUT) in the Universal Dependencies parsing framework, covering a range of genres.  We focus on
the problem of prepositional phrase attachment: whether the head of a prepositional phrase attaches
to a verb or to some other dependent of the verb. We manually selected a small set of sentences from
these treebanks that had potentially ambiguous attachments.

\paragraph{Reasoning about perspectives} (PERSPECTRUM) Given a debate-worthy natural language claim,
the task is to identify the set of relevant argumentative sentences that represent perspectives
for/against the claim~\cite{chen2019seeing}. We focus on the stance prediction sub-task: a binary
prediction of whether a relevant perspective is for/against the given claim.

\paragraph{Discrete Reasoning Over Paragraphs} (DROP) A reading comprehension dataset that requires
numerical reasoning, e.g., adding, sorting, and counting numbers in paragraphs~\cite{dua2019drop}.
In order to compute the consistency metric for the span answers of DROP, we report the average
number of contrast sets in which $F_1$ for all instances is above $0.8$.

\paragraph{\quoref} A reading comprehension task with span selection questions that require
coreference resolution~\cite{dasigi2019quoref}. In this dataset, most questions can be localized to
a single event in the passage, and reference an argument in that event that is typically a pronoun
or other anaphoric reference.  Correctly answering the question requires resolving the pronoun. We
use the same definition for consistency for \quoref as we did for \emph{DROP}.

\paragraph{Reasoning Over Paragraph Effects in Situations} (ROPES) A reading comprehension dataset
that requires applying knowledge from a background passage to new
situations~\cite{lin2019reasoning}.  This task has background paragraphs drawn mostly from science
texts that describe causes and effects (e.g., that brightly colored flowers attract insects), and
situations written by crowd workers that instantiate either the cause (e.g., bright colors) or the
effect (e.g., attracting insects).  Questions are written that query the application of the
statements in the background paragraphs to the instantiated situation.  Correctly answering the
questions is intended to require understanding how free-form causal language can be understood and
applied.  We use the same consistency metric for ROPES as we did for DROP and \quoref.

\paragraph{BoolQ} A dataset of reading comprehension instances with Boolean (yes or no)
answers~\cite{clark2019boolq}.  These questions were obtained from organic Google search queries and
paired with paragraphs from Wikipedia pages that are labeled as sufficient to deduce the answer.  As
the questions are drawn from a distribution of what people search for on the internet, there is no
clear set of ``intended phenomena'' in this data; it is an eclectic mix of different kinds of
questions.

\paragraph{MC-TACO} A dataset of reading comprehension questions about multiple temporal
common-sense phenomena~\cite{zhou2019going}.  Given a short paragraph (often a single sentence), a
question, and a collection of candidate answers, the task is to determine which of the candidate
answers are plausible.  For example, the paragraph might describe a storm and the question might ask
how long the storm lasted, with candidate answers ranging from seconds to weeks.  This dataset is
intended to test a system's knowledge of typical event durations, orderings, and frequency.  As the
paragraph does not contain the information necessary to answer the question, this dataset is largely
a test of background (common sense) knowledge.

\section{Contrast Set Details}\label{appendix:contrast-set-details}

\subsection{NLVR2}
\paragraph{Text Perturbation Strategies} We use the following text perturbation strategies for NLVR2:
\begin{itemize}[nosep,leftmargin=6mm]
  \item Perturbing quantifiers, e.g., \emph{There is at least one dog} $\to$ \emph{There is exactly
    one dog}.
  \item Perturbing numbers, e.g., \emph{There is at least one dog} $\to$ \emph{There are at least
    two dogs}.
  \item Perturbing entities, e.g., \emph{There is at least one dog} $\to$ \emph{There is at least
    one cat}.
  \item Perturbing properties of entities, e.g., \emph{There is at least one yellow dog} $\to$
    \emph{There is at least one green dog}.
\end{itemize}

\paragraph{Image Perturbation Strategies} For image perturbations, the annotators collected images
that are perceptually and/or conceptually close to the hypothesized decision boundary, i.e., they
represent a minimal change in some concrete aspect of the image. For example, for an image pair with
2 dogs on the left and 1 dog on the right and the sentence \emph{There are more dogs on the left than
the right}, a reasonable image change would be to replace the right-hand image with an image of
two dogs.

\paragraph{Model} We use LXMERT~\cite{lxmert} trained on the NLVR2 training dataset.

\paragraph{Contrast Set Statistics} Five annotators created 983 perturbed instances that form 479
contrast sets. Annotation took approximately thirty seconds per textual perturbation and two minutes
per image perturbation.

\subsection{IMDb}

\paragraph{Perturbation Strategies} 

We minimally perturb reviews to flip the label while ensuring that the review remains coherent and
factually consistent. Here, we provide example revisions:

\begin{framed}
            \small 
            \noindent
            \textbf{Original (Negative):} I had quite high hopes for this film, even though it got a bad review in the paper. I was extremely {\color{red}tolerant}, and sat through the entire film. I felt quite {\color{red}sick} by the end. \\ 
            \small 
            \noindent
            \textbf{New (Positive):} I had quite high hopes for this film, even though it got a bad review in the paper. I was extremely {\color{blue} amused}, and sat through the entire film. I felt quite {\color{blue} happy} by the end. \\
            \small 
            \noindent
            \textbf{Original (Positive):} This is the \textcolor{red}{greatest} film I saw in 2002, whereas I'm used to mainstream movies. It is \textcolor{red}{rich and makes a} \textcolor{red}{beautiful artistic act} from these 11 short films. From the technical info (the chosen directors), I feared it would have an anti-American basis, but ... it's a kind of (11 times) \textcolor{red}{personal tribute}. \textcolor{red}{The weakest point} comes from Y. Chahine : he does not manage to ``swallow his pride'' and considers this event as a well-merited punishment ... It is \textcolor{red}{really the weakest} part of the movie, but this testifies of a real freedom of speech for the whole piece.  \\
            \small 
            \noindent
            \textbf{New (Negative):} This is the \textcolor{blue}{most horrendous} film I saw in 2002, whereas I'm used to mainstream movies. It is \textcolor{blue}{low budgeted and makes a less than beautiful artistic act} from these 11 short films. From the technical info (the chosen directors), I feared it would have an anti-American basis, but ... it's a kind of (11 times) \textcolor{blue}{the same}. \textcolor{blue}{One of the weakest point} comes from Y. Chahine : he does not manage to ``swallow his pride'' and considers this event as a well-merited punishment ... It is \textcolor{blue}{not the weakest} part of the movie, but this testifies of a real freedom of speech for the whole piece.
        \end{framed}

\paragraph{Model} We use the same BERT model setup and training data as \citet{kaushik2019learning}
which allows us to fairly compare the crowd and expert revisions.

\paragraph{Contrast Set Statistics} We use 100 reviews from the validation set and 488 from the test
set of \citet{kaushik2019learning}. Three annotators used approximately 70 hours to construct and
validate the dataset.

\subsection{MATRES}
\label{subsec:matres}
MATRES has three sections: TimeBank, AQUAINT, and Platinum, with the Platinum section serving as the
test set. We use 239 instances (30\% of the dataset) from Platinum.

\paragraph{Perturbation Strategies} The annotators perturb one or more of the following aspects:
appearance order in text, tense of verb(s), and temporal conjunction words. Below are example
revisions:~\smallskip
\begin{itemize}[nosep,leftmargin=6mm]
    \small
    \item Colonel Collins \textbf{followed} a normal progression once she was \textbf{picked} as a
      NASA astronaut. (original sentence: ``followed'' is after ``picked'')
    \item Once Colonel Collins was \textbf{picked} as a NASA astronaut, she \textbf{followed} a
      normal progression. (appearance order change in text; ``followed'' is still after ``picked'')
    \item Colonel Collins \textbf{followed} a normal progression before she was \textbf{picked} as a
      NASA astronaut. (changed the temporal conjunction word from ``once'' to ``before'' and
      ``followed'' is now before ``picked'')
    \item Volleyball is a popular sport in the area, and more than 200 people were \textbf{watching}
      the game, the chief \textbf{said}. (original sentence: ``watching'' is before ``said'')
    \item Volleyball is a popular sport in the area, and more than 200 people would be
      \textbf{watching} the game, the chief \textbf{said}. (changed the verb tense: ``watching'' is
      after ``said'')
\end{itemize}

\paragraph{Model} We use CogCompTime 2.0~\cite{NingSuRo19}.

\paragraph{Contrast Set Statistics} Two annotators created 401 perturbed instances that form 239
contrast sets. The annotators used approximately 25 hours to construct and validate the dataset.

\paragraph{Analysis} We recorded the perturbation strategy used for each example. 49\% of the
perturbations changed the ``appearance order'', 31\% changed the ``tense'', 24\% changed the
``temporal conjunction words'', and 10\% had other changes. We double count the examples that have
multiple perturbations. The model accuracy on the different perturbations is reported in the table
below. 

\begin{table}[h]
\centering
\footnotesize{
\begin{tabular}{ll}
\toprule
\textbf{Perturbation Type} & \textbf{Accuracy} \\
\midrule
Overall & 63.3\% \\
\midrule 
Appearance Order & 66.5\% \\
Tense Change & 61.8\% \\
Temporal Conjunction & 60.0\% \\
Other Changes & 61.8\% \\
\bottomrule
\end{tabular}
}
\caption{Accuracy breakdown of the perturbation types for MATRES.}
\label{tab:matres_breakdown}
\end{table}

\subsection{Syntactic Parsing}

\paragraph{Perturbation Strategies} The annotators perturbed noun phrases adjacent to prepositions
(leaving the preposition unchanged). For example, \emph{The clerics demanded talks with local US
commanders} $\to$ \emph{The clerics demanded talks with great urgency}. The different semantic
content of the noun phrase changes the syntactic path from the preposition \textit{with} to the
parent word of the parent of the preposition; in the initial example, the parent is
\textit{commanders} and the grandparent is the noun \textit{talks}; in the perturbed version, the
grandparent is now the verb \textit{demanded}.

\paragraph{Model} We use a biaffine parser following the architecture of \citet{dozat2016deep} with ELMo embeddings~\cite{PetersELMo2018},
trained on the combination of the training sets for the treebanks that we drew test examples from
(GUM, EWT, LinES, and ParTUT).

\paragraph{Contrast Set Statistics} One annotator created 150 perturbed examples that form 150
contrast sets. 75 of the contrast sets consist of a sentence in which a prepositional phrase
attaches to a verb, paired with an altered version where it attaches to a noun instead. The other 75
sentences were altered in the opposite direction.

\paragraph{Analysis} The process of creating a perturbation for a syntactic parse is highly
time-consuming. Only a small fraction of sentences in the test set could be altered in the desired
way, even after filtering to find relevant syntactic structures and eliminate unambiguous
prepositions (e.g. \textit{of} always attaches to a noun modifying a noun, making it impossible to
change the attachment without changing the preposition). Further, once a potentially ambiguous
sentence was identified, annotators had to come up with an alternative noun phrase that sounded
natural and did not require extensive changes to the structure of the sentence. They then had to
re-annotate the relevant section of the sentence, which could include new POS tags, new UD word
features, and new arc labels. On average, each perturbation took 10--15 minutes. Expanding the scope
of this augmented dataset to cover other syntactic features, such as adjective scope, apposition
versus conjunction, and other forms of clausal attachment, would allow for a significantly larger
dataset but would require a large amount of annotator time. The very poor contrast consistency on
our dataset (17.3\%) suggests that this would be a worthwhile investment to create a more rigorous
parsing evaluation.

Notably, the model's accuracy for predicting the target prepositions' grandparents in the original,
unaltered tree (64.7\%) is significantly lower than the model's accuracy for grandparents of all
words (78.41\%) and for grandparents of all prepositions (78.95\%) in the original data. This
indicates that these structures are already difficult for the parser due to structural ambiguity.

\subsection{PERSPECTRUM}

\paragraph{Perturbation Strategies} The annotators perturbed examples in multiple steps. First, they
created non-trivial negations of the claim, e.g., \emph{Should we live in space?} $\to$ \emph{Should we
drop the ambition to live in space?}. Next, they labeled the perturbed claim with respect to each
perspective. For example:
\begin{framed}
    \small
    \noindent
    \textbf{Claim:} Should we {\color{red}live} in space? \\ 
    \textbf{Perspective:} Humanity in many ways defines itself through exploration and space is the next logical frontier. \\ 
    \textbf{Label:} True \\
    \noindent\rule{\textwidth}{1pt}
    \textbf{Claim:} Should we {\color{blue}drop the ambition to live} in space? \\ 
    \textbf{Perspective:} Humanity in many ways defines itself through exploration and space is the next logical frontier. \\ 
    \textbf{Label:} False
\end{framed}

\paragraph{Model} We use a \textsc{RoBERTa} model~\cite{liu2019roberta} finetuned on PERSPECTRUM
following the training process from \cite{chen2019seeing}.

\paragraph{Contrast Set Statistics} The annotators created 217 perturbed instances that form 217
contrast sets. Each example took approximately three minutes to annotate: one minute for an
annotator to negate each claim and one minute each for two separate annotators to adjudicate stance
labels for each contrastive claim-perspective pair.

\subsection{DROP}

\paragraph{Perturbation Strategies} See Section~\ref{sec:examples} in the main text for details
about our perturbation strategies.

\paragraph{Model} We use MTMSN~\cite{hu2019mtmsn}, a DROP question answering model that is built on
top of BERT Large~\cite{devlin2018BERT}.

\paragraph{Contrast Set Statistics} The total size of the augmented test set is 947 examples and
contains a total of 623 contrast sets. Three annotators used approximately 16  hours to construct
and validate the dataset.

\paragraph{Analysis} We bucket $100$ of the perturbed instances into the three categories of
perturbations described in Section~\ref{sec:examples}. For each subset, we evaluate MTMSN's
performance and show the results in the Table below. 

\begin{table}[h]
\centering
\footnotesize{
\begin{tabular}{lll}
\toprule
\textbf{Perturbation Type} & \textbf{Frequency} & \textbf{Accuracy} \\
\midrule 
Adding Compositional Steps & 38\% & 67.5 $F_1$ \\
Inversion of Semantics & 37\% & 53.2 $F_1$ \\
Re-ordering Events & 25\% & 47.3 $F_1$ \\
\bottomrule
\end{tabular}
}
\caption{Accuracy breakdown of the perturbation types for DROP.}
\label{tab:drop_breakdown}
\end{table}

\subsection{\quoref}

\paragraph{Perturbation Strategies} We use the following perturbation strategies for \quoref:
\begin{itemize}[nosep,leftmargin=6mm]
    \item Perturb questions whose answers are entities to instead make the answers a property of
      those entities, e.g., \emph{Who hides their identity ...} $\to$ \emph{What is the nationality
      of the person who hides their identity ...}.
    \item Perturb questions to add compositionality, e.g., \emph{What is the name of the person ...}
      $\to$ \emph{What is the name of the father of the person ...}.
    \item Add sentences between referring expressions and antecedents to the context paragraphs.
    \item Replace antecedents with less frequent named entities of the same type in the context
      paragraphs.
\end{itemize}

\paragraph{Model} We use XLNet-QA, the best model from \citet{dasigi2019quoref}, which is a span
extraction model built on top of XLNet~\cite{yang2019xlnet}.

\paragraph{Contrast Set Statistics} Four annotators created 700 instances that form 415 contrast
sets. The mean contrast set size (including the original example) is $2.7 (\pm 1.2)$. The annotators
used approximately 35 hours to construct and validate the dataset.

\subsection{\ropes}
\paragraph{Perturbation Strategies} We use the following perturbation strategies for \ropes:
\begin{itemize}[nosep,leftmargin=6mm]
    \item Perturbing the background to have the opposite causes and effects or qualitative relation,
      e.g., \emph{Gibberellins are hormones that cause the plant to grow} $\to$ \emph{Gibberellins
      are hormones that cause the plant to stop growing.}
    \item Perturbing the situation to associate different entities with different instantiations of
      a certain cause or effect. For example, \emph{Grey tree frogs live in wooded areas and are
      difficult to see when on tree trunks. Green tree frogs live in wetlands with lots of grass and
      tall plants.} $\to$ \emph{Grey tree frogs live in wetlands areas and are difficult to see when
      on stormy days in the plants. Green tree frogs live in wetlands with lots of leaves to hide
      on.}
    \item Perturbing the situation to have more complex reasoning steps, e.g., \emph{Sue put 2 cubes
      of sugar into her tea. Ann decided to use granulated sugar and added the same amount of sugar
      to her tea.} $\to$ \emph{Sue has 2 cubes of sugar but Ann has the same amount of granulated
      sugar.  They exchange the sugar to each other and put the sugar to their ice tea.}
    \item Perturbing the questions to have presuppositions that match the situation and background.
\end{itemize}
    
\paragraph{Model} We use the best model from \citet{lin2019reasoning}, which is a span extraction
model built on top of a RoBERTa model~\cite{liu2019roberta} that is first finetuned on
RACE~\cite{lai2017race}.

\paragraph{Contrast Set Statistics} Two annotators created 974 perturbed instances which form 974
contrast sets. The annotators used approximately 65 hours to construct and validate the dataset.

\subsection{BoolQ} 

\paragraph{Perturbation Strategies} We use a diverse set of perturbations, including adjective,
entity, and event changes. We show three representative examples below:
        \begin{framed}
            \small 
            \noindent
            \textbf{Paragraph:} The Fate of the Furious premiered in Berlin on April 4, 2017, and was theatrically released in the United States on April 14, 2017, playing in 3D, IMAX 3D and 4DX internationally\ldots A spinoff film starring Johnson and Statham's characters is scheduled for release in August 2019, while the ninth and tenth films are scheduled for releases on the years 2020 and 2021. \\ 
            \textbf{Question:} Is ``Fate and the Furious'' the {\color{red}last movie}? \\
            \textbf{Answer:} False \\
            \textbf{New Question:} Is ``Fate and the Furious'' {\color{blue} the first of multiple movies}? \\
            \textbf{New Answer:} True \\
            \textbf{Perturbation Strategy:} Adjective Change 
        \end{framed}
        \begin{framed}
            \small 
            \noindent
            \textbf{Paragraph:} Sanders played football primarily at cornerback, but also as a kick returner, punt returner, and occasionally wide receiver\ldots An outfielder in baseball, he played professionally for the New York Yankees, the Atlanta Braves, the Cincinnati Reds and the San Francisco Giants, and participated in the 1992 World Series with the Braves.\\ 
            \textbf{Question:} Did Deion Sanders ever {\color{red}win} a world series? \\
            \textbf{Answer:} False \\
            \textbf{New Question:} Did Deion Sanders ever {\color{blue}play in} a world series? \\ 
            \textbf{New Answer:} True \\
            \textbf{Perturbation strategy:} Event Change 
        \end{framed}
        \begin{framed}
            \noindent
            \small 
            \textbf{Paragraph:} The White House is the official residence and workplace of the President of the United States. It is located at 1600 Pennsylvania Avenue NW in Washington, D.C. and has been the residence of every U.S. President since John Adams in 1800. The term is often used as a metonym for the president and his advisers. \\ 
            \textbf{Question:} {\color{red}Does the president} live in the White House? \\
            \textbf{Answer:} True \\ 
            \textbf{New Question:} {\color{blue}Did George Washington} live in the White House? \\
            \textbf{New Answer:} False \\ 
            \textbf{Perturbation Strategy:} Entity Change  
        \end{framed}
        
\paragraph{Model} We use \textsc{RoBERTa} base and follow the standard finetuning process
from~\citet{liu2019roberta}. 

\paragraph{Contrast Set Statistics} The annotators created 339 perturbed questions generated that
form 70 contrast sets. One annotator created the dataset and a separate annotator verified it. This
entire process took approximately 16 hours.

\subsection{MC-TACO}

\paragraph{Perturbation Strategies} The main goal when perturbing MC-TACO questions is to retain a
similar question that requires the same temporal knowledge to answer, while there are additional
constraints with slightly different related context that changes the answers. We also modified the
answers accordingly to make sure the question has a combination of plausible and implausible
candidates.

\paragraph{Model} We use the best baseline model from the original paper~\cite{zhou2019going} which
is based on $\textsc{RoBERTa}_{base}$ \cite{liu2019roberta}.

\paragraph{Contrast Set Statistics} The annotators created 646 perturbed question-answer pairs that
form 646 contrast sets.  Two annotators used approximately 12 hours to construct and validate the
dataset.

\end{document}